\documentclass{article}

     \PassOptionsToPackage{numbers, compress}{natbib}
     \usepackage[preprint]{neurips_2019}

\usepackage[utf8]{inputenc} % allow utf-8 input
\usepackage[T1]{fontenc}    % use 8-bit T1 fonts
\usepackage{hyperref}       % hyperlinks
\usepackage{url}            % simple URL typesetting
\usepackage{booktabs}       % professional-quality tables
\usepackage{amsfonts}       % blackboard math symbols
\usepackage{nicefrac}       % compact symbols for 1/2, etc.
\usepackage{microtype}      % microtypography

\usepackage{gensymb}
\usepackage{url,amssymb,amsmath,amsthm,amscd,paralist,bbm,wrapfig}
\usepackage{tkz-graph}
\usepackage{tikz-cd}
\usepackage{amsmath, bm}
\usepackage{fancyhdr}
\usepackage{amssymb}
\usepackage{graphicx}
\usepackage{algorithm}
\usepackage{algorithmic} %//format of the algorithm 

\usepackage{mathtools}

\newcommand{\R}{\mathbb{R}}

\newcommand{\al}{\alpha}

\newcommand{\A}{\pmb{A}}
\newcommand{\x}{\pmb{x}}
\newcommand{\y}{\pmb{y}}
\newcommand{\n}{\pmb{\eta}}
\newcommand{\z}{\pmb{z}}
\newcommand{\bftheta}{\pmb{\theta}}

\newcommand{\G}{\mathcal{G}}

\title{Low Shot Learning with Untrained Neural Networks for Imaging Inverse Problems}

\author{%
  Oscar Leong\thanks{This work was performed at the Data Science Summer Institute (DSSI) under the auspices of the U.S. Department of Energy by Lawrence Livermore National Laboratory under Contract DE-AC52-07NA27344.} \\
  Rice University\\
  \texttt{oscar.f.leong@rice.edu} \\
  \And
  Wesam Sakla \\
  Lawrence Livermore National Laboratory \\
  \texttt{sakla1@llnl.gov} \\
}

\begin{document}

\maketitle

\begin{abstract}
  Employing deep neural networks as natural image priors to solve inverse problems either requires large amounts of data to sufficiently train expressive generative models or can succeed with no data via untrained neural networks. However, very few works have considered how to interpolate between these no- to high-data regimes. In particular, how can one use the availability of a small amount of data (even $5-25$ examples) to one's advantage in solving these inverse problems and can a system's performance increase as the amount of data increases as well? In this work, we consider solving linear inverse problems when given a small number of examples of images that are drawn from the same distribution as the image of interest. Comparing to untrained neural networks that use no data, we show how one can pre-train a neural network with a few given examples to improve reconstruction results in compressed sensing and semantic image recovery problems such as colorization. Our approach leads to improved reconstruction as the amount of available data increases and is on par with fully trained generative models, while requiring less than $1 \%$ of the data needed to train a generative model.
\end{abstract}

\section{Introduction}

We study the problem of recovering an image $\x_0 \in \R^n$ from $m$ linear measurements of the form \[\y_0 = \A \x_0 + \n \in \R^m\] where $\A \in \R^{m \times n}$ is a known measurement operator and $\n \in \R^m$ denotes the noise in our system. Problems of this form are ubiquitous in various domains ranging from image processing, machine learning, and computer vision. Typically, the problem's difficulty is a result of its ill-posedness due to the underdetermined nature of the system. To resolve this ambiguity, many approaches enforce that the image must obey a natural image model. While traditional approaches typically use hand-crafted priors such as sparsity in the wavelet basis \cite{Dono2007}, recent approaches inspired by deep learning to create such natural image model surrogates have shown to outperform these methods.

\paragraph{Deep Generative Priors:} Advancements in generative modelling have allowed for deep neural networks to create highly realistic samples from a number of complex natural image classes. Popular generative models to use as natural image priors are latent variable models such as Generative Adversarial Networks (GANs) \cite{Goodfellow2014} and Variational Autoencoders (VAEs) \cite{Kingma2014}. This is in large part due to the fact that they provide a low-dimensional parameterization of the natural image manifold that can be directly exploited in inverse imaging tasks. When enforced as a natural image prior, these models have shown to outperform traditional methods and provide theoretical guarantees in problems such as compressed sensing \cite{Boraetal2017, Wuetal2019, Hand2017, Huangetal2018, SH2018, Hussein2019}, phase retrieval \cite{Leong2018, Shamshad2018, Hyderetal2019}, and blind deconvolution/demodulation \cite{Asimetal2018, HJ2019}. However, there are two main drawbacks of using deep generative models as natural image priors. The first is that they require a large amount of data to train, e.g., hundreds of thousands of images to generate novel celebrity faces. Additionally, they suffer from a non-trivial representation error due to the fact that they model the natural image manifold through a low-dimensional parameterization.

\paragraph{Untrained Neural Network Priors:} On the opposite end of the data spectrum, recent works have shown that randomly initialized neural networks can act as natural image priors without any learning. \cite{Ulyanov2017} first showed this to be the case by solving tasks such as denoising, inpainting, and super-resolution via optimizing over the parameters of a convolutional neural network to fit to a single image. The results showed that the neural network exhibited a bias towards natural images, but due to the high overparameterization in the network, required early stopping to succeed. A simpler model was later introduced in \cite{HH2019} which was, in fact, underparameterized and was able to both compress images while solving various linear inverse problems. Both methods require no training data and do not suffer from the same representation error as generative models do. Similar to generative models, they have shown to be successful image priors in a variety of inverse problems \cite{HH2019, Heckel2019, VVetal2019, JH2019}.

Based on these two approaches, we would like to investigate how can one interpolate between these data regimes in a way that improves upon work with untrained neural network priors and ultimately reaches or exceeds the success of generative priors. More specifically, we would like to develop an algorithm that 1) performs just as well as untrained neural networks with no data and 2) improves performance as the amount of provided data increases.

\paragraph{Our contributions:}

We introduce a framework to solve inverse problems given a few examples (e.g., $5-25$) drawn from the same data distribution as the image of interest (e.g., if the true image is of a human face, the examples are also human faces). Our main contributions are the following:

\begin{itemize}
    \item We show how one can pre-train a neural network using a few examples drawn from the data distribution of the image of interest. Inspired by \cite{Bojanowski2017}, we propose to jointly learn a latent space and parameters of the network to fit to the examples that are given and compare the use of an $\ell_2$ reconstruction loss and a kernel-based Maximum Mean Discrepancy (MMD) loss. 
    \item We then propose to solve the inverse problem via a two-step process. We first optimize over the pre-trained network's latent space. Once a solution is found, we then refine our estimate by optimizing over the latent space and parameters of the network jointly to improve our solution. \cite{Hussein2019} found this method to work well in the case when the network is a fully trained generative model, and we show here that even a pre-trained neural network from a small number of examples can benefit from such an approach.
    
    \item We show that our approach improves upon untrained neural networks in compressed sensing even with as few as $5$ examples from the data distribution and exhibits improvements as the number of examples increases. We also show that semantics can be learned from these few examples in problems such as colorization where untrained neural networks fail. With only $100$ examples, our model's performance is competitive with fully trained generative models.
\end{itemize}

\paragraph{Related work:} We mention that there has been previous work \cite{VVetal2019} in investigating how to use a small amount of data to help solve the compressed sensing problem. The authors use an untrained neural network as a natural image prior and, when given a small amount of data, adopt a learned regularization term when solving the inverse problem. This term is derived by posing the recovery problem as a Maximum a Posteriori (MAP) estimation problem and by placing a Gaussian prior on the weights of the untrained network. While we have not compared our method to this learned regularization approach here, we aim to do so in a subsequent manuscript.

\section{Low Shot Learning For Imaging Inverse Problems}
\label{low_shot_sec}

We consider the problem of recovering an image $\x_0 \in \R^n$ from noisy linear measurements of the form $\y_0 = \A\x_0 + \n \in \R^m$ where $\A \in \R^{m \times n}$ and $m \leqslant n$. We also assume that $\x_0$ is drawn from a particular data distribution $\mathcal{D}$ and that we are given a low number of examples drawn from the same distribution, i.e., $\x_0 \sim \mathcal{D}$ and we are given $\x_i \sim \mathcal{D}$ where $i \in [S]$. Here and throughout this work, we refer to these examples drawn from $\mathcal{D}$ as \textit{low shots}.

 We propose using the range of a deep neural network as a natural image model. In particular, we model the image $\x_0$ as the output of a neural network $\G(\z; \bftheta)$, where $\z \in \R^k$ is a latent code and $\bftheta \in \R^P$ are the parameters of the network. 

\paragraph{Pre-training:} Prior to solving the inverse problem, we propose to first pre-train the network using the low shots that are given. More specifically, we fit the weights and input of the neural network to the low shots to provide a crude approximation to the data distribution underlying the image of interest. Given low shots $\{\x_i\}_{i=1}^S$, we aim to find latent codes $\{\z_i\}_{i=1}^S$ and parameters $\bftheta$ that solve \begin{align}
    \min_{\bftheta,\z_1,\dots,\z_S} \frac{1}{S} \sum_{i=1}^S \mathcal{L}(\mathcal{G}(\z_i; \bftheta), \x_i). \label{general_training_objective}
    \end{align} where $\mathcal{L} : \R^n \times \R^n \rightarrow \R$ is a loss function. We investigate the use of different loss functions in a later section. The resulting optimal parameters found are denoted by $\hat{\bftheta}$, $\hat{\z}_1,\dots,\hat{\z}_S$.

\paragraph{Solving the inverse problem:} Using the weights found via pre-training, we begin solving the inverse problem by first optimizing over the latent code space to find an approximate solution: \begin{align}
    \min_{\z} \frac{1}{2} \Big\| \A\mathcal{G}(\z;\hat{\bftheta}) - \y_0 \Big\|^2_2. \label{solve_wrt_z}
\end{align} We investigated different ways to initialize the latent code and found that sampling from a multivariate Gaussian distribution fit using $\{\hat{\z}_i\}_{i=1}^S$ was sufficient. Note here that we keep the parameters of the network fixed after training. The intuition is that we want to use the semantics regarding the data distribution learned via pre-training the network's parameters and find the optimal latent code that corresponds to the image of interest. Once the optimal latent code $\hat{\z}$ is found, we then refine our solution by solving \begin{align}
    \min_{\bftheta,\z} \frac{1}{2} \Big\| \A\mathcal{G}(\z;\bftheta) - \y_0 \Big\|^2_2 \label{solve_wrt_z_and_theta}
\end{align} with $\hat{\bftheta}$ and $\hat{\z}$ as our initial iterates. The resulting parameters $\bftheta_0$ and $\z_0$ give our final estimate:  $\x_0 \approx \mathcal{G}(\z_0; \bftheta_0)$.

\paragraph{Losses to learn the data distribution:} We discuss two loss functions that we considered in our experiments to learn semantics regarding the underlying data distribution. The first is a simple $\ell_2$ reconstruction loss to promote data fidelity, i.e., we pre-train our network by solving \begin{align}
    \min_{\bftheta,\z_1,\dots,\z_S}  \frac{1}{S}\sum_{i=1}^S \left\|\mathcal{G}(\z_i; \bftheta) - \x_i\right\|_2^2. \label{ell2_training_objective}
\end{align} While \cite{Bojanowski2017} used a combination of the Laplacian-L1 loss and $\ell_2$ loss, we found the $\ell_2$ loss to work well.

The second loss is an estimate of the kernel MMD for comparing two probability distributions using only finitely many samples \cite{Gretton2012}. In our case, given a kernel $k(\cdot,\cdot)$\footnote{We only consider the Gaussian kernel $k_{\al}(\x_1, \x_2) := \exp(-\|\x_1 - \x_2\|_2^2 / \al)$ in our experiments.} and low shots $\x_j \sim \mathcal{D}$ for $j \in [S]$, we want to find parameters and $S$ inputs that solve the following: \begin{align}
    \min_{\bftheta,\z_1,\dots,\z_S} \frac{1}{\binom{S}{2}}\sum_{i \neq i'} k(\mathcal{G}(\z_i; \bftheta), \mathcal{G}(\z_{i'}; \bftheta)) + \frac{1}{\binom{S}{2}}\sum_{j \neq j'} k(\x_j, \x_{j'}) - \frac{2}{\binom{S}{2}} \sum_{i \neq j} k(\mathcal{G}(\z_i; \bftheta), \x_j). \label{MMD_training_objective}
\end{align} 

We compare the success of these two loss functions in the following section.

\section{Experiments}

We now consider solving inverse problems with our approach and compare to three different baselines: an untrained neural network, optimizing the latent space of a trained Wasserstein GAN \cite{WGAN2017} with gradient penalty \cite{WGANGP2017}, and the image-adaptivity approach of \cite{Hussein2019} (IAGAN). Each method uses the same DCGAN architecture with a latent code dimension of $128$. In each problem, the image of interest is from a hold-out test set from the CelebA dataset \cite{CelebA}. The GAN was trained on a corpus of over $200,000$ $64 \times 64$ celebrity images and our low-shot models were trained on small ($5-100$ images) subsets of this. 

\paragraph{Implementation details:} For training our low shot models, we used the Adam optimizer \cite{ADAM} with a learning rate of $10^{-3}$ for $50,000$ iterations. To solve the inverse problem, the latent space of the GAN was optimized using Adam for $1700$ iterations and a learning rate of $10^{-1}$. For IAGAN, the parameters and latent code were then jointly optimized for an additional $350$ iterations with a learning rate of $10^{-4}$. Our low shot models followed a similar procedure where we first optimized the latent space for $1250$ iterations with a learning rate of $5*10^{-2}$. Then the parameters and latent space were jointly optimized for $350$ iterations with a learning rate of $10^{-4}$. For the untrained neural network, we solely optimized the network's parameters using the RMSProp optimizer \cite{RMSProp} with a learning rate of $10^{-3}$ and momentum of $0.9$ as in \cite{VVetal2019}. For low compression ratios (defined below) in compressed sensing ($\frac{m}{n} \leqslant 0.025$), we used $350$ iterations as we noticed overfitting in this noisy setting. When $0.025 < \frac{m}{n} \leqslant 0.5$ and $\frac{m}{n} > 0.5$, we used $500$ and $1000$ iterations, respectively.

\paragraph{Compressed Sensing:} We first consider the compressed sensing problem where we want to recover an image $\x_0 \in \R^n$ from random Gaussian measurements of the form $\y_0 = \A \x_0 \in \R^m$ where $\A \in \R^{m \times n}$ has i.i.d. $\mathcal{N}(0,1)$ entries with $m \ll n$. We refer to amount of undersampling $\frac{m}{n}$ as the \textit{compression ratio}. We trained our models using the two different loss functions proposed in the previous section for various numbers of shots $S \in [5,10,15,25,50, 100]$. 

\begin{figure}[!htbp]
    \centering
    \includegraphics[width=\textwidth]{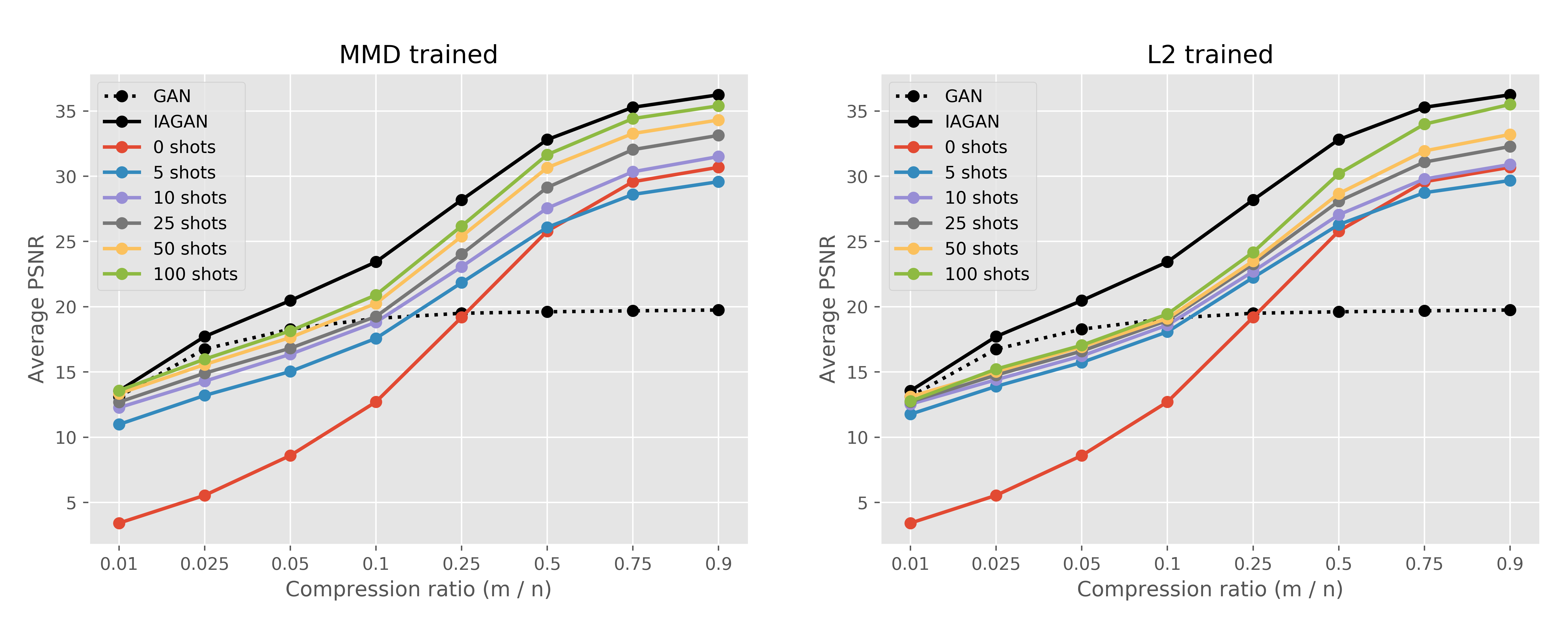}
    \caption{Average PSNR over $50$ different test images for models trained with the MMD loss (left) and the $\ell_2$ loss (right).}
    \label{fig:comparisons_cs}
\end{figure}

Figure \ref{fig:comparisons_cs} compares the average PSNR for each method at various compression ratios over $50$ different test images and different loss functions. We note that as the number of shots increases, our method continues to improve and we see comparable performance between our method with only $100$ shots and optimizing over the latent code space of a fully trained GAN. While the $\ell_2$ trained nets perform slightly better than the MMD trained nets for low numbers of shots, the MMD trained nets improve more steadily and consistently as the number of shots increases. While we expect IAGAN to be superior due to being trained with over $200,000$ images, the MMD trained model's performance with $100$ images is not far behind. We note that for higher numbers of measurements, the untrained neural network's performance surpasses that of our $5$ shot models. This is mainly due to the fact that we optimized the untrained neural network's parameters for a longer period of time and with a higher learning rate than our low shot models in this easier setting.

\begin{figure}[!htbp]
    \centering
    \includegraphics[width = \textwidth]{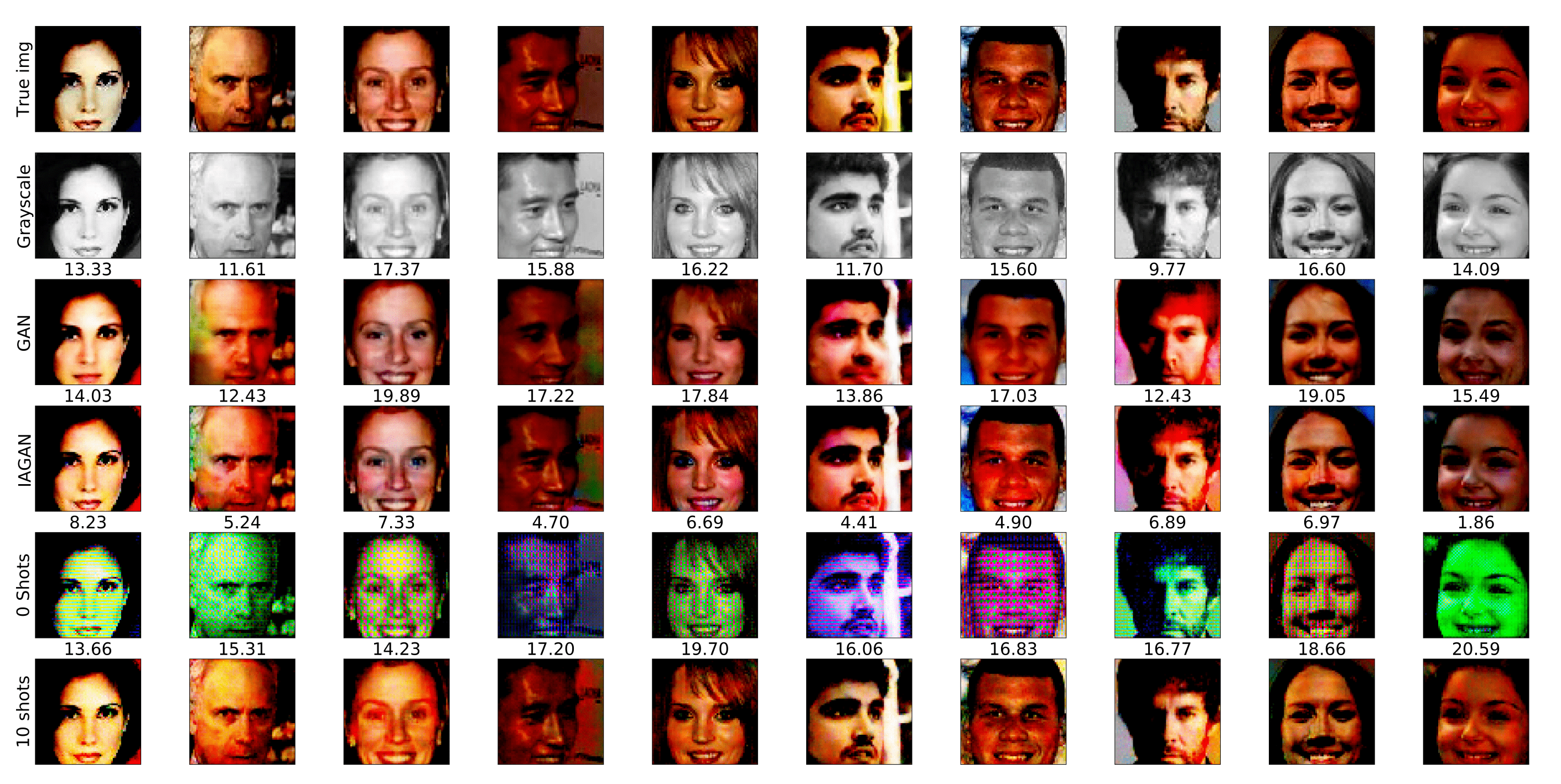}
    \caption{Qualitative results for the colorization task. Rows 1 and 2 are the true and grayscale images while rows 3-6 are the GAN, IAGAN, untrained neural network, and our $10$ shot model's reconstructions, respectively. The PSNR of each result is shown above the reconstruction.}
    \label{fig:comparisons_colorization}
\end{figure}

\paragraph{Colorization:} We now consider an inverse problem that requires an understanding of the underlying data's semantics: the colorization task. Here we want to recover an RGB image $\x_0 \in \R^{64 \times 64 \times 3}$ from its grayscale version $\y_0 = \A \x_0 \in \R^{64 \times 64}$. The operator $\A$ mixes the color channels of the image via the ITU-R 601-2 luma transform (the same transform used by the Python Imaging Library (PIL)). Untrained neural networks clearly fail in solving this type of problem since they have no prior information regarding the data distribution. We compare our model trained with $10$ shots using the MMD loss to the various baselines in Figure \ref{fig:comparisons_colorization}. Note that even with only $10$ previous examples, our model does not fall prey to the usual issues with using untrained neural networks for colorization. Our algorithm provides faithful image reconstructions that are even on par with a trained GAN.

\subsubsection*{Disclaimer}

This document was prepared as an account of work sponsored by an agency of the United States government. Neither the United States government nor Lawrence Livermore National Security, LLC, nor any of their employees makes any warranty, expressed or implied, or assumes any legal liability or responsibility for the accuracy, completeness, or usefulness of any information, apparatus, product, or process disclosed, or represents that its use would not infringe privately owned rights. Reference herein to any specific commercial product, process, or service by trade name, trademark, manufacturer, or otherwise does not necessarily constitute or imply its endorsement, recommendation, or favoring by the United States government or Lawrence Livermore National Security, LLC. The views and opinions of authors expressed herein do not necessarily state or reflect those of the United States government or Lawrence Livermore National Security, LLC, and shall not be used for advertising or product endorsement purposes.

%\subsubsection*{Acknowledgments}

\bibliographystyle{plain}

\bibliography{low_shot.bib}

\newpage

%For training our low shot models, we used the Adam optimizer \cite{ADAM} with a learning rate of $10^{-3}$ for $50,000$ iterations. The latent space of the GAN was optimized using Adam for $1700$ iterations and a learning rate of $10^{-1}$. For IAGAN, the parameters and latent code were then jointly optimized for an additional $350$ iterations with a learning rate of $10^{-4}$. Our low shot models followed a similar procedure where we first optimized the latent space for $1250$ iterations with a learning rate of $5*10^{-2}$. Then the parameters and latent space were jointly optimized for $350$ iterations with a learning rate of $10^{-4}$. For the untrained neural network, we solely optimized the network's parameters using the RMSProp optimizer \cite{RMSProp} with a learning rate of $10^{-3}$ and momentum of $0.9$. For low compression ratios $\frac{m}{n}$, we used $350$ iterations as we noticed overfitting in this noisy setting. For higher numbers of measurements we used $500$ and $1000$ iterations.

\end{document}